\documentclass[sigconf]{acmart}
\usepackage{physics}
\usepackage{hyperref}
\usepackage{url}

\NewDocumentCommand{\codeword}{v}{%
\texttt{#1}%
}

\AtBeginDocument{%
  \providecommand\BibTeX{{%
    \normalfont B\kern-0.5em{\scshape i\kern-0.25em b}\kern-0.8em\TeX}}}

\setcopyright{acmcopyright}
\copyrightyear{2018}
\acmYear{2018}
\acmDOI{XXXXXXX.XXXXXXX}

\acmConference[ICSEW'24]{46th International Conference on Software Engineering Workshops}{April 14--20,
  2024}{Lisbon, Portugal}
\acmPrice{15.00}
\acmISBN{978-1-4503-XXXX-X/18/06}

\begin{document}

\title{\chatforquant: A Chatbot for Quantum}


\author{Yaiza Aragon\'es-Soria}
\orcid{0000-0002-8880-8957}
\affiliation{%
  \institution{Constructor Institute Schaffhausen}
  \streetaddress{Rheinweg 9}
  \city{Schaffhausen}
  \country{Switzerland}
  \postcode{43017-6221}
}
\email{yaiza.aragonessoria@constructor.org}

\author{Manuel Oriol}
\orcid{0000-0003-4069-7626}
\affiliation{%
  \institution{Constructor Institute Schaffhausen}
  \streetaddress{Rheinweg 9}
  \city{Schaffhausen}
  \country{Switzerland}
  \postcode{43017-6221}
}
\affiliation{%
    \institution{Constructor University Bremen}
    \streetaddress{Campus Ring 1}
    \city{Bremen}
    \country{Germany}
    \postcode{28759}
}
\email{mo@constructor.org}


%the name of the tool (ideally has a free domain) 
\newcommand{\chatforquant}{C4Q}

\begin{abstract}
Quantum computing is a growing field that promises many real-world applications such as quantum cryptography or quantum finance. 
The number of people able to use quantum computing is however still very small.
This limitation comes from the difficulty to understand the concepts and to know how to start coding. 
Therefore, there is a need for tools that can assist non-expert in overcoming this complexity. 
One possibility would be to use existing conversational agents.
Unfortunately ChatGPT and other Large-Language Models produce inaccurate results.

This article presents \chatforquant, a chatbot that answers accurately basic questions and guides users when trying to code quantum programs. 
Contrary to other approaches \chatforquant~ uses a pre-trained large language model only to discover and classify user requests. 
It then generates an accurate answer using an own engine.
Thanks to this architectural design, \chatforquant's answers are always correct, and thus \chatforquant~ can become a support tool that makes quantum computing more available to non-experts.

%Thanks to this architectural design, \chatforquant~ answers are always correct, since \chatforquant~ prioritizes abstaining from answering rather than providing inaccurate information.

\end{abstract}




\received{December 7, 2023}
\received[accepted]{January 11, 2024}
\received[revised]{January 25, 2024}

\maketitle

\section{Introduction}
% Motivation
%Story line: 

%coding quantum gates isn't for everyone

%we use LLM to help users to understand and code quantum

%ChatGPT is not able to give correct answers about quantum computation.

%It makes no sense to answer questions about logic using probabilities.

%make quantum computing more available to software developers from students to researchers.

Quantum computing promises to bring large performance improvements for some classes of problems.
Most developers willing to learn how to code for quantum computers need to follow a course and buy books that explain the intricacies of coding with qubits and quantum gates.
Though qubits and quantum gates have already much capabilities, they are not intuitive and easy to use by regular developers.

Unfortunately, using ChatGPT~\cite{chatGPT} or other generic Large-Language Models (LLMs) reports a mix of correct and blatantly wrong answers with perfect assurance.
% WHY WE BUILT C4Q
In a previous study~\cite{our_paper}, ChatGPT's answers to basic questions on quantum computing revealed severe inconsistencies. 
%ChatGPT's responses range from correct to entirely incorrect, and sometimes included incomplete or erroneous information.
Authors asked ChatGPT to give a definition and a circuit representation of Pauli Z as well as to apply Pauli Z on $\ket{1}$.
Although the answer contained a circuit representation which could be considered correct, the definition as well as the application of Pauli Z were wrong (Fig.~\ref{f:chatgpt_answer}).

This article presents \chatforquant,\footnote{\url{http://chatbot4quantum.org}} a solution that helps developers acquire and verify their knowledge on quantum computing by using a chatbot specialized for quantum computing.
\chatforquant~ answers basic questions about qubits and quantum gates in an accurate manner.
%, it also suggests algorithms to use for solving more complicated problems.

This article is divided as follows. Section~\ref{s:motivational_example} presents the tool \chatforquant.
Section~\ref{s:architecture} describes the architecture of \chatforquant. 
Section~\ref{s:example-of-interaction} illustrates with an example the interaction dynamics between \chatforquant~ and a user.
Section~\ref{s:results} evaluates \chatforquant.
Section~\ref{s:rw} presents related work and Section~\ref{s:conc} concludes and presents future work.

\section{A Chatbot for Quantum} \label{s:motivational_example}
\begin{figure}[htbp]
  \centering
  \includegraphics[width=\linewidth]{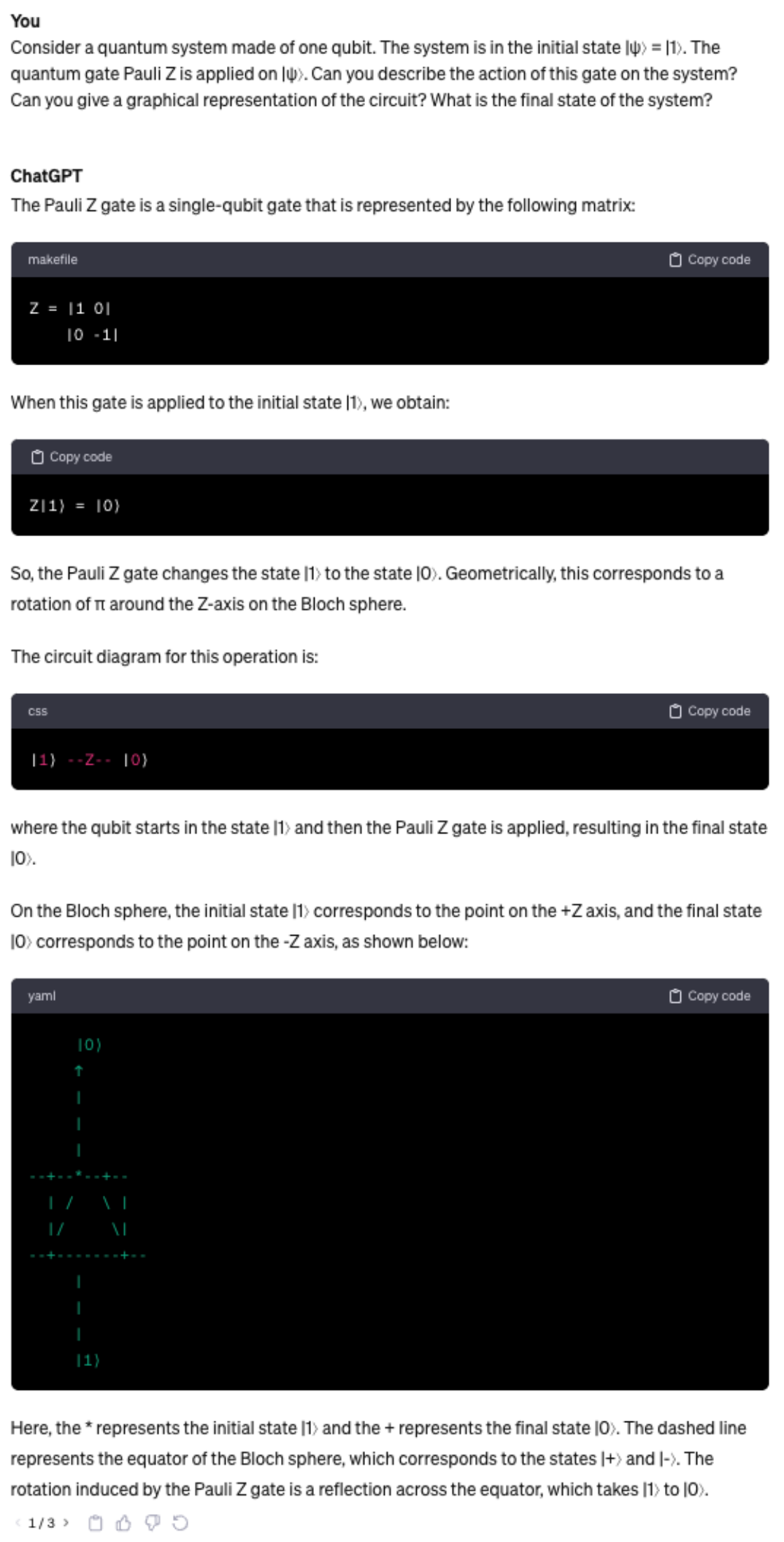}
  \caption{Answer of ChatGPT when asked about the quantum gate Pauli Z.}
  \label{f:chatgpt_answer}
  \Description{}
\end{figure}

This section presents \chatforquant: a new chatbot aimed at helping developers with understanding and coding quantum programs. Currently, \chatforquant~ focuses on the fundamentals of quantum computing, with the intention of broadening its functionality to encompass more advanced concepts in the future.
% EXPLANATION C4Q
\chatforquant~ is a solution to the shortcomings of ChatGPT~\cite{our_paper}. 
\chatforquant~ is a chatbot that specializes in addressing enquiries related to quantum computing, particularly focusing on a defined set of quantum gates. 
This set encompasses fundamental gates such as the identity, the Pauli gates, the $S$ and $S^\dag$ gates, the Hadamard gate, phase gates, rotations, the CNOT gate, CZ gate, and the SWAP gate.

For each quantum gate within this set, \chatforquant~ is able to fulfill three distinct types of queries:
\begin{enumerate}
    \item \emph{Defining} the quantum gate. This involves providing a comprehensive description of the specific quantum gate, elucidating its fundamental properties and characteristics.
    \item \emph{Drawing} the quantum gate. \chatforquant~ is equipped to generate a circuit representation of the quantum gate.
    \item \emph{Applying} the gate on an initial state. Users can inquire about the practical application of the quantum gate on an initial state, and \chatforquant~ computes the resulting state.
\end{enumerate}

Before generating an answer, \chatforquant~ analyzes the user's question and requests confirmation to ensure accurate understanding. This approach serves two purposes for \chatforquant: it verifies the absence of any misinterpretation of the question and utilizes the user's input to continually enhance the training of the LLM.

\begin{figure}[htbp]
  \centering
  \includegraphics[width=\linewidth]{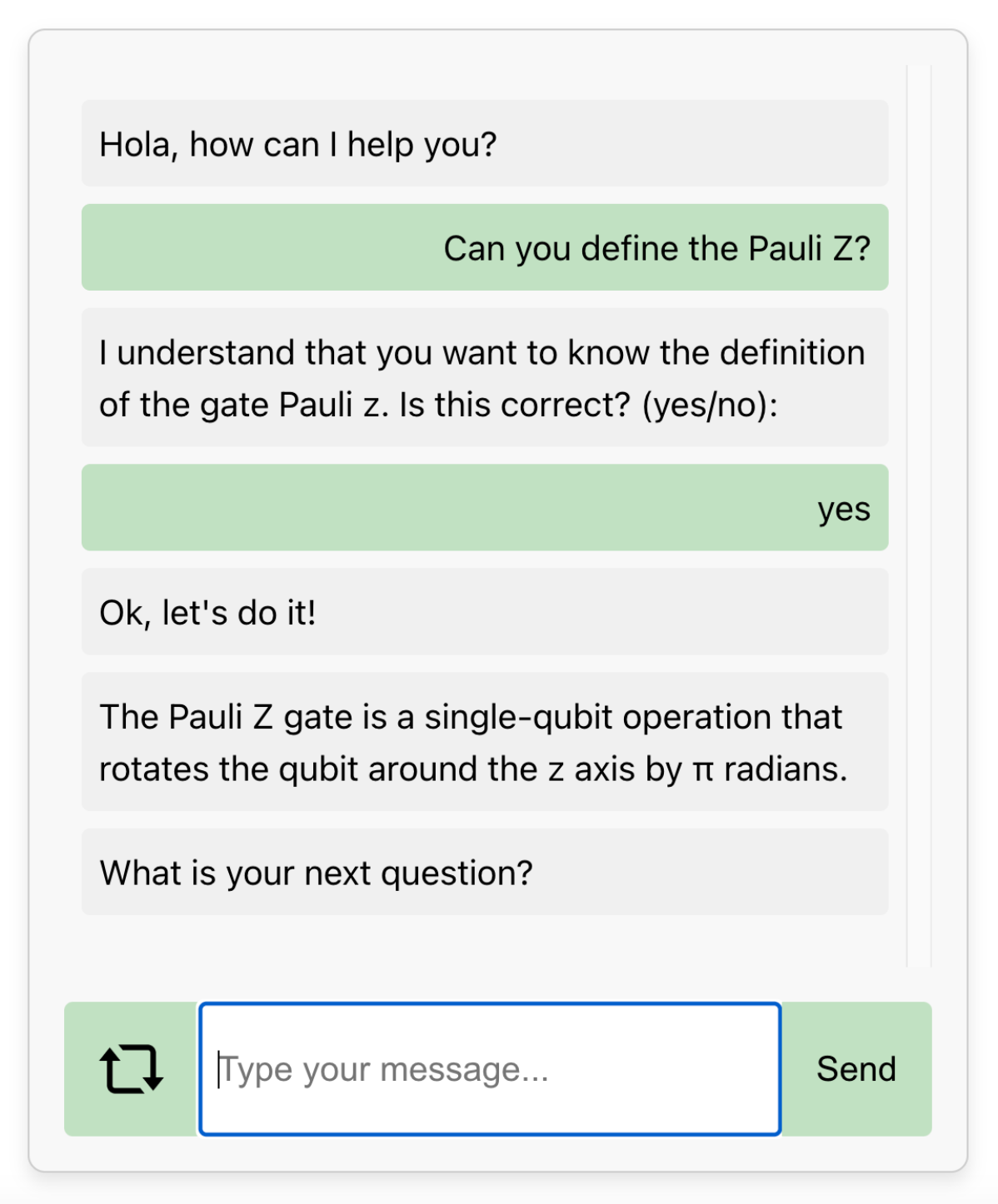}
  \caption{Answer of \chatforquant~ when asked to define the quantum gate Pauli Z.}
  \label{f:define_pauliZ}
  \Description{}
\end{figure}

\begin{figure}[htbp]
  \centering
  \includegraphics[width=\linewidth]{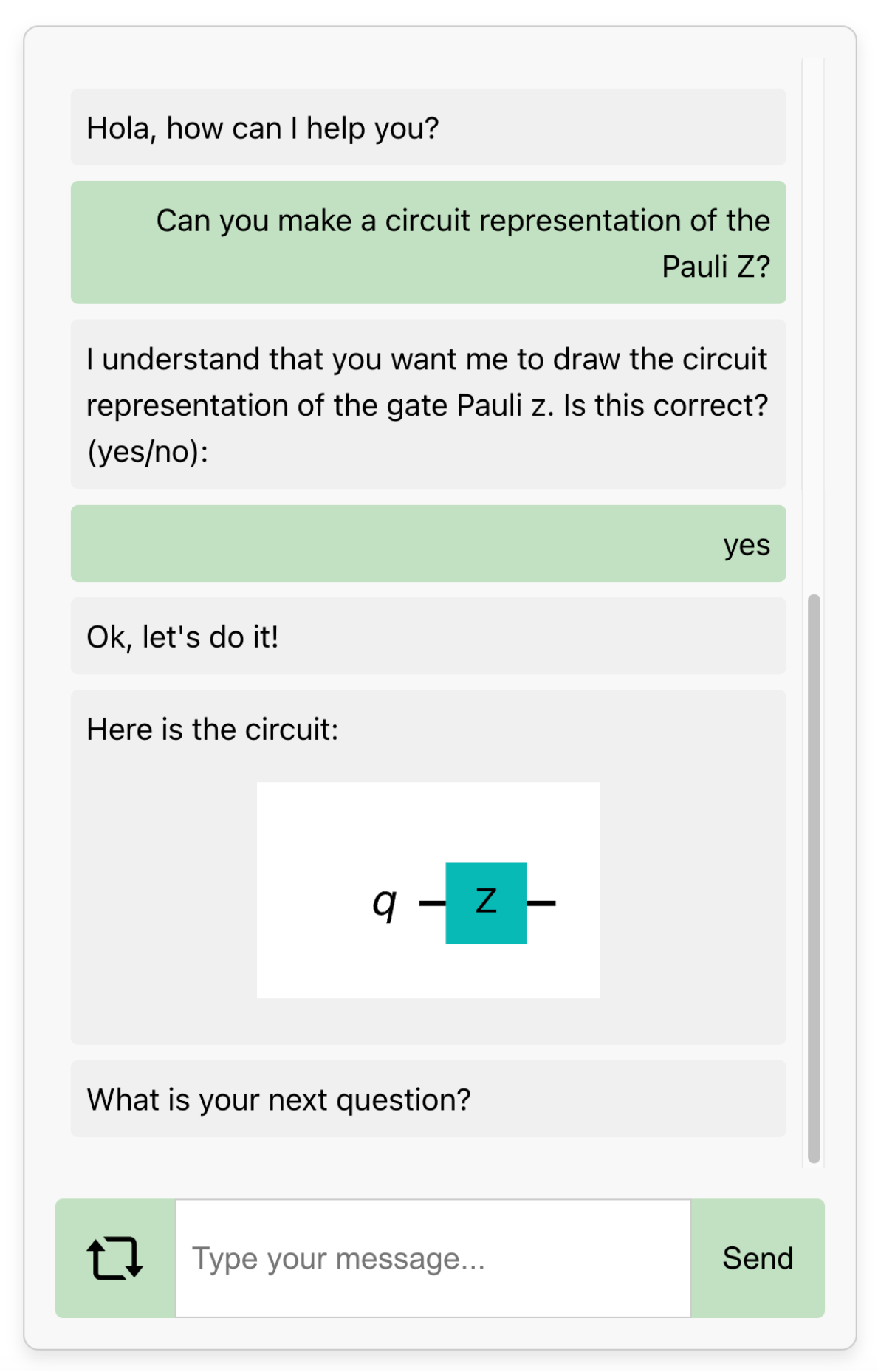}
  \caption{Answer of \chatforquant~ when asked to create a circuit representation of the quantum gate Pauli Z.}
  \label{f:draw_pauliZ}
  \Description{}
\end{figure}

\begin{figure}[htbp]
  \centering
  \includegraphics[width=\linewidth]{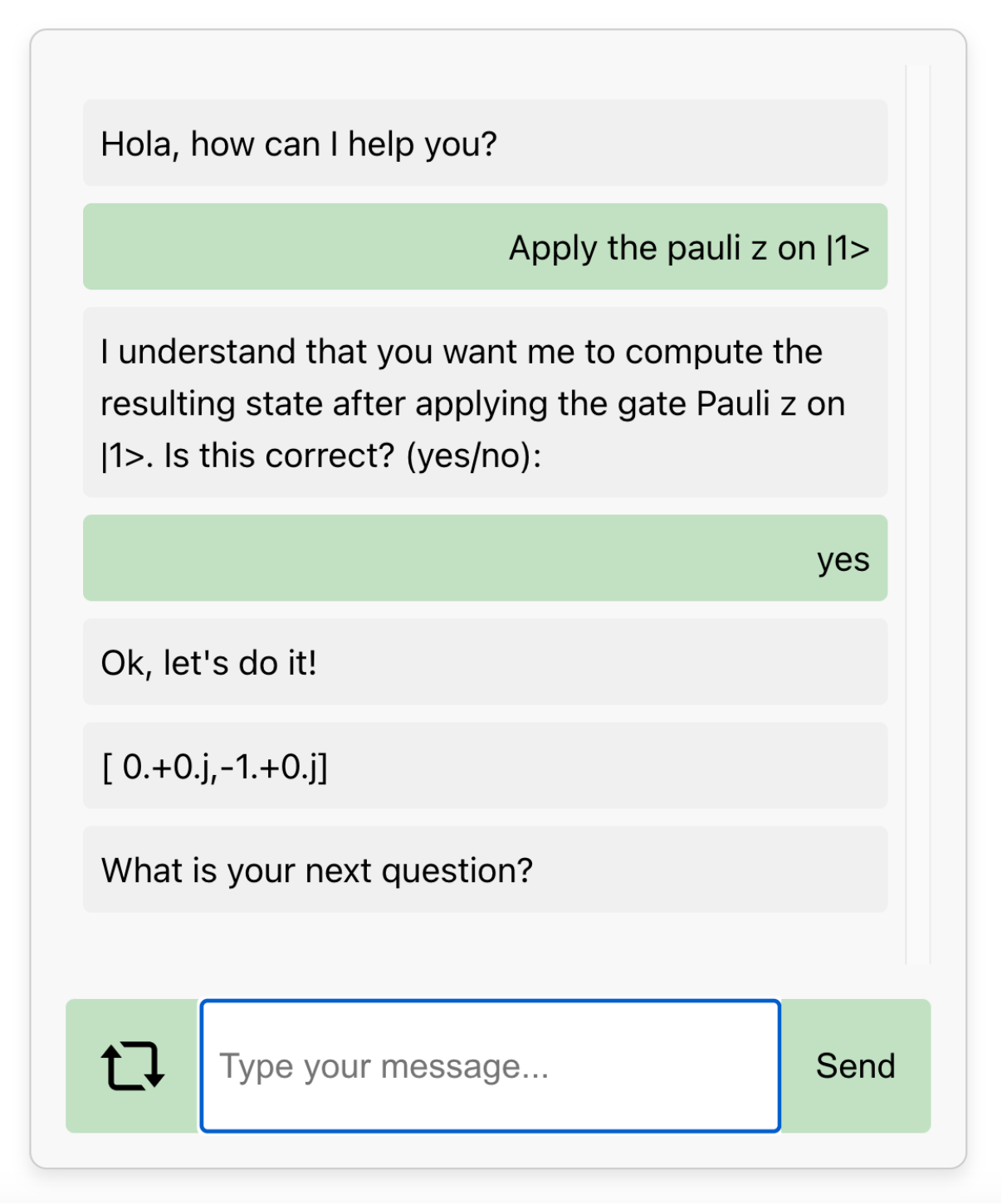}
  \caption{Answer of \chatforquant~ when asked to apply the quantum gate Pauli Z on $\ket{1}$.}
  \label{f:apply_pauliZ}
  \Description{}
\end{figure}

% EXAMPLE OF C4Q
In contrast to ChatGPT, \chatforquant~ serves as a reliable solution by consistently delivering correct answers. 
Comparing to the previous example on ChatGPT (Fig.~\ref{f:chatgpt_answer}), \chatforquant~ accurately handles it in three separate steps $-$ defining, drawing and applying the Pauli Z (Figures \ref{f:define_pauliZ}, \ref{f:draw_pauliZ} and \ref{f:apply_pauliZ}). 
The next section explains the architecture of \chatforquant, which is the key for \chatforquant's accuracy.
%This persistent accuracy is further detailed in the next section, where we explain the architecture of C4Q.

\section{Architecture of \chatforquant} \label{s:architecture}
This section presents the architecture of \chatforquant. 
It starts with a broad overview of different modules integrated in \chatforquant.
Then, it discusses the implementation specifics of each module.

\begin{figure*}[htbp]
  \centering
  \includegraphics[width=.8\linewidth]{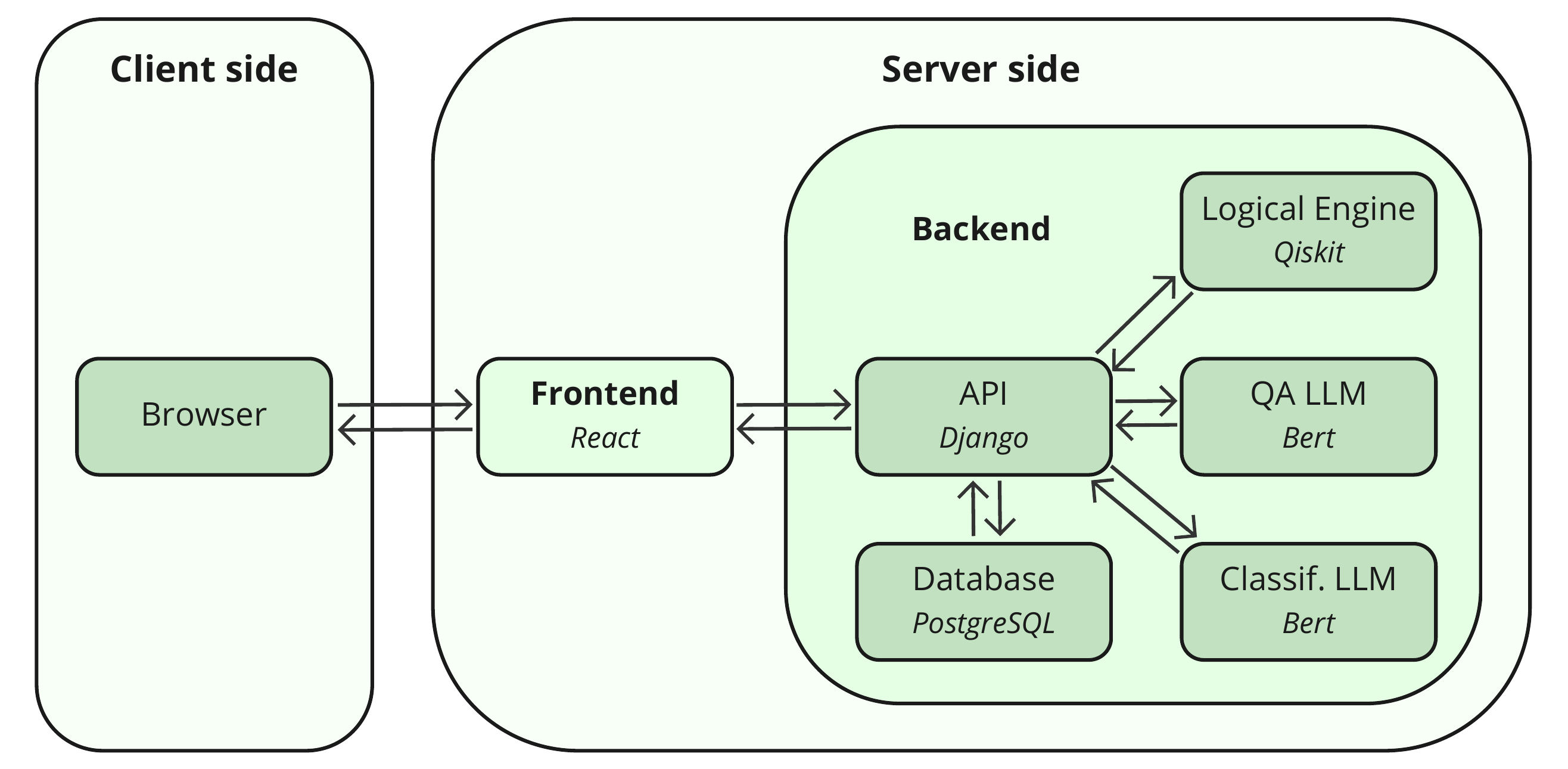}
  \caption{Architecture diagram of C4Q.}
  \label{f:architecture_diagram}
  \Description{}
\end{figure*}

\chatforquant~ is made of several modules that take care of different parts of the requests (see Figure~\ref{f:architecture_diagram}).
The frontend is developed using React, ensuring a simple and user-friendly interface. 
In the backend, five key modules contribute to \chatforquant's operation:

\begin{enumerate}
    \item \emph{API}. This module serves as the interface, handling communication between the frontend and the various backend components.
    \item \emph{Database}. This module stores the user information as well as the messages that the user exchanges with the chatbot.
    \item \emph{Classification LLM}. Leveraging a fine-tuned model for classification, this module classifies user questions into three categories: define, draw or apply a gate.
    \item \emph{Question Answering (QA) LLM}. This module uses a fine-tuned QA bert model and enables \chatforquant~ to extract information from user questions.
    \item \emph{Logical engine}. Using the classification and the information extracted by the LLMs, this module generates an accurate answer.
\end{enumerate}

The following sections describe each module in more details.

\subsection{API}
This section provides a comprehensive overview of the API module, a component serving as the interface between various backend modules and the frontend.

The primary objective of the API module is to act as a seamless bridge, ensuring effective communication among different backend components and establishing a connection between backend and frontend. 
This includes managing the database, transmitting information to the frontend, interacting with both the classification LLM and the QA LLM, and requesting the logical engine to generate accurate responses.

The API is built using the Django framework and currently incorporates two models:
\begin{enumerate}
    \item \emph{User Model}. This model manages user-related information and facilitates user-specific interactions.
    \item \emph{Message model}. This model handles message-related data, including the parameters required by the logical engine for generating responses.
\end{enumerate}

In future versions of \chatforquant, the API module aims to introduce a conversation model. This addition will encompass the bundling of messages, enabling users to archive and store entire conversations. We believe that this extension will enhance the user experience by providing a structured and organized approach to managing interactions within the system.

\subsection{Database}

The Database module operates by leveraging PostgreSQL~\cite{postgresql} with the psycopg2~\cite{psycopg2} adapter to establish and manage connections seamlessly. 
The choice of PostgreSQL, coupled with the adept handling provided by the psycopg2 adapter, underscores the commitment to a robust and scalable database solution within the system.

The database stores the conversation between the user and \chatforquant. Upon the conclusion of a session, the conversation is deleted in order to keep a clean database. \chatforquant~ only keeps the user questions that are correctly classified, enabling future improvements of \chatforquant's classification LLM.

\subsection{Classification LLM}
The classification-LLM module sorts user questions so that the logical engine can generate an accurate answer.

According to \chatforquant's abilities mentioned in Section~\ref{s:motivational_example}, the classification LLM classifies user questions in three categories: defining a quantum gate, drawing a quantum gate and applying a quantum gate. 
To achieve this, we used the pretrained model BertForSequenceClassification \cite{bert_for_classification}, specifically in its \verb|bert-base-uncased| version, as the basis for a fine-tuning process. 

To ensure an effective fine-tuning, we carefully curated a tailored dataset to simulate real-world user queries associated with the three question categories. 
Each dataset entry was labeled with the accurate classification, establishing a solid groundwork for training the LLM.

Following the recommendations~\cite{bertpaper}, the fine-tuning process adhered to the following specific parameters:
\begin{itemize}
\item A batch size of 16 tokens and 4 epochs were chosen, striking a balance between computational efficiency and model convergence. 
\item The generated dataset underwent a thorough shuffling, and an 80-20 data split was implemented, allocating 80\% for training and 20\% for validation. 
\item The optimization process was executed using the AdamW optimizer~\cite{AdamW, AdamWpaper}, with a learning rate, $lr$, of $2\cdot10^{-5}$ and an epsilon, $eps$,  of $1\cdot10^{-8}$ for numerical stability.
\end{itemize}

\subsection{Question Answering LLM}
This section provides an in-depth exploration of thq QA LLM, the backend module that extracts information from user questions. 
Its core functionality is to extract the necessary parameters for the logical engine to generate accurate answers. 
The QA LLM is specifically tailored to obtain the phase shift for questions related to phase gates and to identify the axis and angle for queries about rotations.\newline

Developing the QA-LLM module involved using simpletransformers~\cite{simpletransformers}.
The starting point was the pretrained bert model called QuestionAnsweringModel~\cite{bert_for_qa}, specifically the ``bert-base-uncased'' variant.

The fine-tuning of the model uses two distinct datasets.
In the dataset for phase gates, each entry contains a contextual question about a phase gate, the specific question the LLM should learn to answer, i.e., ``what is the phase shift?'', and the corresponding answer derived from the provided context.

Similarly, the dataset centered around rotations included entries with contextual questions about rotations, the two questions the LLM should learn to answer, i.e., ``what is the angle of the rotation?'' and ``what is the axis of the rotation?'', and the corresponding answers aligned with the context.

For the fine-tuning process, a train and evaluation batch size of 16 tokens were employed over a span of four epochs. 
The generated data, shuffled and divided, was allocated in a 50-50 split, with 50\% dedicated to training and the remaining 50\% reserved for validation. 
The learning rate, $lr$, was set to $2\cdot10^{-5}$ and an epsilon, $eps$, to $1\cdot10^{-8}$.

These steps in dataset creation and fine-tuning contribute to the QA LLM's ability to precisely extract the pertinent parameters from user queries, enhancing the \chatforquant's responsiveness in addressing inquiries related to phase gates and rotations.

\subsection{Logical Engine}
The Logical Engine is the backend module designed to produce the required logic for responding to user questions.

The API sends the information extracted by the LLMs from the user question to the Logical Engine.
%The Logical Engine receives the information extracted by the LLMs from the user question via the API. 
Using this information, and by seamlessly integrating with Qiskit, the Logical Engine ensures the precise execution of quantum computations. 
This contributes to the overall responsiveness and accuracy of \chatforquant~ in addressing user questions related to quantum computing. 
The symbiotic relationship between information extraction, logical computation, and quantum handling enables the sophistication and effectiveness of \chatforquant, ensuring a consistent level of accuracy. 
The high accuracy distinguishes \chatforquant~ from other chatbots relying primarily on LLMs and extensive datasets, and thus answering according to probabilities. 

It is important to acknowledge that attention to accuracy has an impact on responsiveness. In comparison to other chatbots like ChatGPT, the conversational experience with \chatforquant~ still lacks a certain human-like quality. 
This is a feature that we plan to improve.

The specific procedures for executing each task – defining, drawing, and applying a quantum gate – are as follows. Each quantum gate is associated with an object containing attributes such as name, Qiskit name, and definition, as well as methods such as define, draw, and apply. The define method outputs a hard coded definition, whereas the draw and apply methods use Qiskit operations to generate the desired output, which depends on parameters such as the initial state.

We have recently become aware of two initiatives that are worth of our attention and deeper comparison to \chatforquant. 
The first is Copilot in Azure Quantum, developed by Microsoft, in the frame of their learning platform. 
It consists in a ``quantum-focused chatbot that helps you write code and better understand quantum concepts'' leveraging Q\#. 
Copilot is built upon OpenAI's GPT-4, and thus it answers using probability, an approach different than \chatforquant. 
The second initiative is Quantum AI Chatbot, a custom version of ChatGPT that claims to be a Quantum Theory Translator and Progressive Trainer with Web Scraping. 
This chatbot is obviously also built upon OpenAI's GPT-4. 
At the moment we have not studied the quality of either system, but we plan to evaluate them in depth in future works.

\subsection{Generation of training data}

To train and test LLMs effectively, the generation of good and representative data is imperative. 
The present approach relies on a meticulously crafted own dataset, simulating how diverse users might pose questions that \chatforquant~ is designed to address.
The aim is to ensure that \chatforquant~ could comprehend user intent regardless of the specific wording.

Initially, we formulated questions in a manner we found natural. 
Based on these questions, we then asked ChatGPT to generate alternative formulations for each question, ensuring a diversity of expressions.

For example, when seeking the definition of a quantum gate, ChatGPT generated a set of 20 distinct queries, such as (\{quantum\_gate\} represents any of the quantum gate):
\begin{itemize}
    \item How would you describe the \{quantum\_gate\}?
    \item Define the \{quantum\_gate\}.
    \item Describe the effect of the \{quantum\_gate\} in a quantum computing system.
    \item What is \{quantum\_gate\}?
\end{itemize}

The procedure to generate data for questions on drawing and applying quantum gates was similar. 
Note that for questions requiring parameters, such as initial state or phase shift, we accounted for both scenarios: one where the user provides the parameters and one where \chatforquant~ uses default values in lack explicit user input. 
This rigorous process ensured a robust and diverse dataset for training and testing \chatforquant.

% - Describe architecture of the chatbot

% - Describe the two engines: LLM and logic engine

% techniques

% models

% data

% - Describe how the engines interact (interface)

% Show a diagram flow

% Explain current activities

% Explain, draw and compute a set of gates

\section{Example of interaction} \label{s:example-of-interaction}
This section uses an example to illustrate the interaction flow between \chatforquant~ and a user. 

\begin{figure*}[htbp]
  \centering
  \includegraphics[width=1\linewidth]{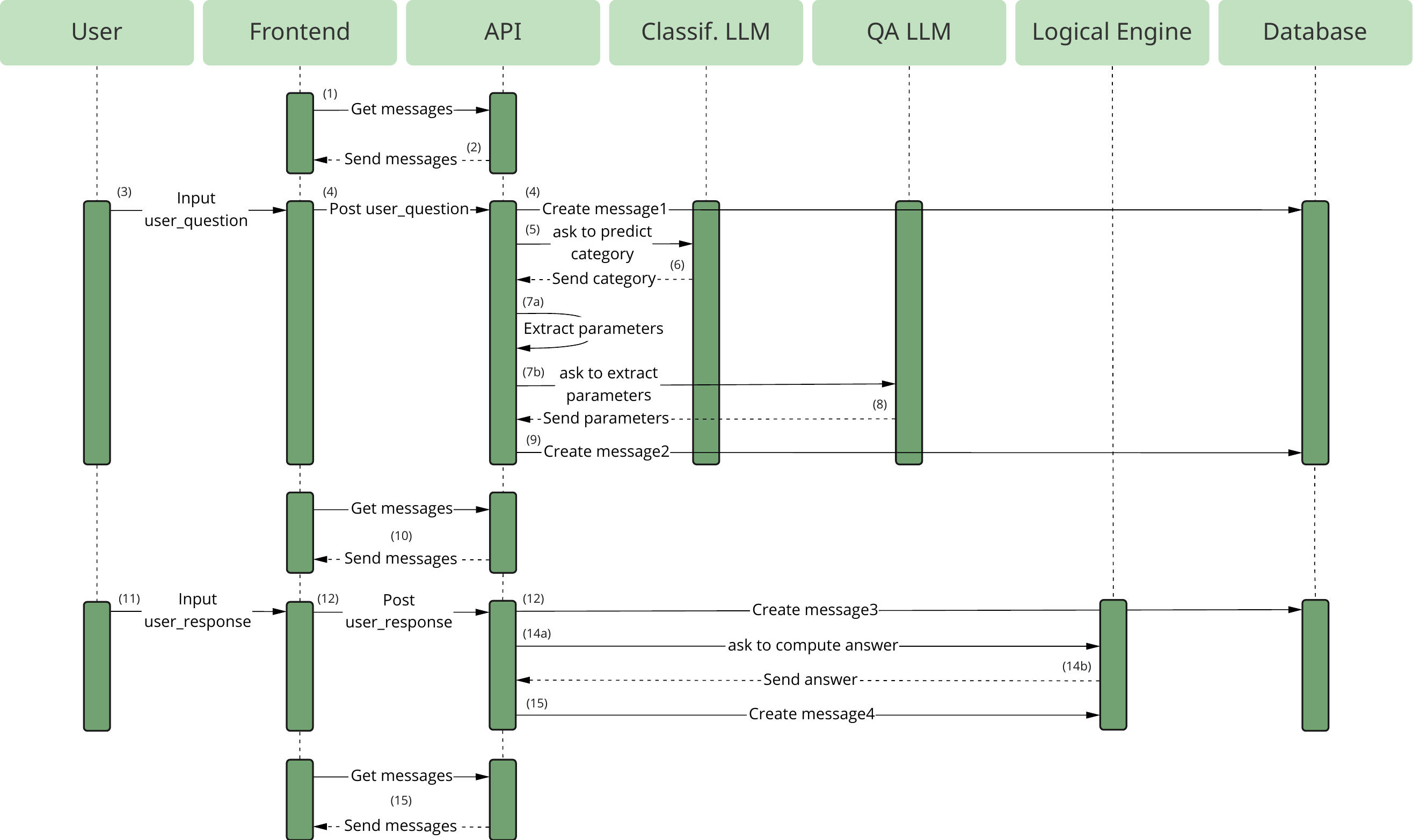}
  \caption{UML Sequence diagram of C4Q. The enumaration corresponds to the sequences described in Section~\ref{s:example-of-interaction}.}
  \label{f:sequence_diagram}
  \Description{}
\end{figure*}

Figure \ref{f:sequence_diagram} is a UML sequence diagram showing a scenario where \chatforquant~ has understood the user's question.
It represents the sequence followed by the request outlined in Figure~\ref{f:apply_pauliZ}:
%The sequence represented in the diagram is the following:

\begin{enumerate}
    \item \emph{Initialization}. The Frontend initiates the conversation by sending a GET request to the API to obtain all available messages, including the greeting message.
    
    \item \emph{Message retrieval}. The API responds by sending all available messages to the Frontend.
    
    \item \emph{User input}. The user inputs a question (\codeword{user_question}).
    
    \item \emph{Message creation}. The Frontend sends a POST request to the API, triggering the creation of a new message (\codeword{message1}) containing the user's question.
    
    \item \emph{Category prediction}. The API requests the Classification LLM to predict the category of the user question.
    
    \item \emph{Category transmission}. The Classification LLM responds with the predicted category.
    
    \item \emph{Parameter extraction}.
    \begin{enumerate}
        \item The API extracts the \codeword{gate_name} and \codeword{initial_state} from the \codeword{user_question}.
        \item The API requests the QA LLM to extract additional parameters from \codeword{user_question}.
    \end{enumerate}

    \item \emph{Parameter transmission}. The QA LLM sends the extracted parameters to the API.

    \item \emph{Message creation}. The API creates \codeword{message2}, with which \chatforquant~ asks the user whether the question \chatforquant~ has understood is correct. The instance \codeword{message2} includes attributes storing \codeword{category} as well as parameters such as \codeword{gate_name} and \codeword{initial_state}.

    \item \emph{Message retrieval}. The Frontend sends a GET request to obtain all messages, and the API responds with all messages, including the confirmation inquiry (\codeword{message2}).

    \item \emph{User input}. The user provides a confirmation (\codeword{user_response = 'yes'}) or rejection (\codeword{user_response = 'no'}) to the question understood by \chatforquant. This example assumes \codeword{user_response = 'yes'} for simplicity.

    \item \emph{Message creation}. The Frontend sends a POST request to the API, triggering the creation of \codeword{message3}, which contains the user's response.

    \item \emph{Data storage}. 
    The API stores \codeword{user_question} and \codeword{category} in a file for future training of the Classification LLM.
    
    \item \emph{Answer computation}. 
        \begin{enumerate}
            \item The API requests the Logical Engine to compute the answer to the user's question.
            \item The Logical Engine computes the answer and sends it back to the API.
        \end{enumerate} 
    
    \item \emph{Message creation}. The API creates \codeword{message4}, which contains the computed answer.

    \item \emph{Message retrieval}. The Frontend sends a GET request to obtain all messages, and the API responds with all messages, including the answer to the user's question.
    
\end{enumerate}

\section{Results and evaluation}\label{s:results}
This section provides evaluation metrics and testing reports for the various modules of the backend.

\subsection{Main results}

The main goal of \chatforquant~ is to overcome deficiencies identified in ChatGPT's responses~\cite{our_paper}.
\chatforquant~ is a chatbot capable of answering queries as well as generating quantum code. 
It has the ability to define, draw, and apply all one- and two-qubit quantum gates that Wikipedia lists as common~\cite{WikipediaQuantumGate}: the identity gate, the Pauli gates, the $S$ and $S^\dag$ gates, the Hadamard gate, phase gates, rotations, the CNOT gate, the CZ gate, and the SWAP gate. 

These gates ensures the creation of a universal gate set, enabling the execution of any operation on a quantum computer.

% \begin{enumerate}
%     \item identity gate
%     \item Pauli gates
%     \item $S$ gate
%     \item $S^\dag$ gate
%     \item Hadamard gate
%     \item phase gate
%     \item rotations
%     \item CNOT gate
%     \item CZ gate
%     \item SWAP gate
% \end{enumerate}
As initial state the user can specify any of the following states: $\ket{0}$, $\ket{1}$, $\ket{+}$, $\ket{-}$, $\ket{r}$, $\ket{l}$, $\ket{00}$, $\ket{01}$, $\ket{10}$, $\ket{11}$, $\ket{\phi^+}$, $\ket{\phi^-}$, $\ket{\psi^+}$, and $\ket{\psi^-}$.
We believe that this represents all states that are used in practice.

\subsection{Code testing}

All modules of \chatforquant's backend are tested with test suites using the pytest framework, with a total of 151 tests including at least one per method. 
The results detailed in the report~\cite{backend_test_report} indicate the successful execution all tests. 

\subsection{LLM evaluation}
\begin{figure}[htbp]
  \centering
  \includegraphics[width=\linewidth]{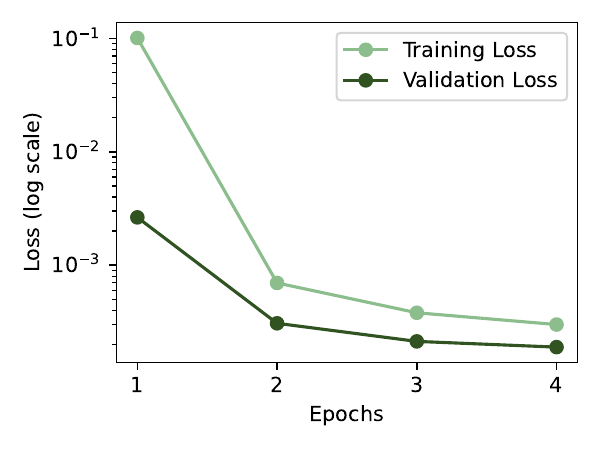}
  \caption{Training and validation loss of the classification LLM.}
  \label{f:training_validation_loss}
  \Description{}
\end{figure}

\begin{figure}[htbp]
  \centering
  \includegraphics[width=\linewidth]{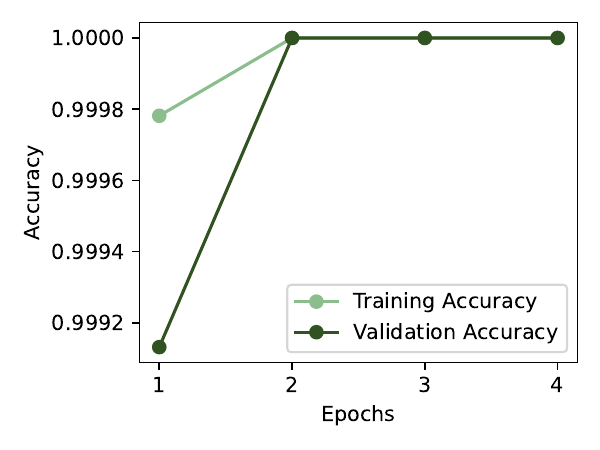}
  \caption{Training and validation accuracy of the classification LLM.}
  \label{f:training_validation_acc}
  \Description{}
\end{figure}

We conducted an evaluation of the performance of both LLMs that are part of \chatforquant. 
The classification LLM was examined by assessing the training and validation loss, along with training and validation accuracy. The evolution of these metrics across epochs is depicted in Figures \ref{f:training_validation_loss} and \ref{f:training_validation_acc}. 
Figure \ref{f:training_validation_loss} shows that both training and validation loss consistently decrease throughout all epochs, with the most substantial decline occurring in the second epoch. 
The training accuracy (see Fig.~\ref{f:training_validation_acc}) as well as the validation accuracy reach their maximum values already by the end of the second epoch.

% \begin{figure}[htbp]
%   \centering
%   \includegraphics[width=\linewidth]{figures/training_validation_plots.pdf}
%   \caption{Training and validation}
%   \label{f:training_validation_plots}
%   \Description{}
% \end{figure}

The performance assessment of the QA LLM consisted in carrying 2000 tests, out of which only 42 failed (see the test report in \cite{QALLM_perfomrance_test_report}). 
These tests show an accuracy of the QA LLM of $0.98$, indicating almost optimal performance. 

In summary, \chatforquant~ effectively responds to basic inquiries about quantum computing and generates corresponding quantum code.
Although the generated quantum code is at the moment hidden from the user, the underlying functionality is robust and aligns with the principles of quantum computation.
Moreover, the performance of \chatforquant~ is backed up by exceptional metrics and a tailored dataset.

%\textcolor{red}{About the class LLM}
%After the completion of the 4 epochs, the model exhibited outstanding performance metrics. The training loss stood at $3\cdot10^{-4}$, indicative of the model's capacity to minimize error during training. The training accuracy reached a perfect score of 1.0, underscoring the LLM's proficiency in learning and categorizing user questions accurately. Furthermore, the validation metrics were equally compelling, with a validation loss of $2\cdot10^{-4}$ and a validation accuracy mirroring the perfection achieved in training. These results affirm the effectiveness of the Classification LLM in precisely classifying user questions and lay the groundwork for its integration into the broader chatbot architecture.

\section{Related Work}\label{s:rw}

Bots are becoming available to software engineers willing to improve their productivity~\citep{8823643,mapping}.
For example, \citet{DBLP:journals/tse/CarrLP17} created a bot that inserts automatically proven contracts in source code, \citet{8115628} made a chat bot that answers questions about APIs, \citet{10.1145/3180155.3180238} made a development assistant able to understand commands for Git and GitHub tasks.

For automated bots generating code, most articles tend to focus on making it as close as possible to what a programmer could have generated. For example, generating automatically patches with explanations~\citep{8823632,10.1145/3183519.3183540} or make refactorings indistinguishable from what a human could have generated~\citep{8823629}.

Many see in ChatGPT, and generally in generative AI, a transforming technology for teaching and research~\citep{murugesan2023, damian2023, yilmaz2023, lim2023}.
Some studies evaluate such as \citep{fisee23Kotovich} whether students can generate fundamental algorithms using ChatGPT3, but none of the systems we tried is providing satisfactory answers. Neither ChatGPT3 nor ChatGPT4 have the ability to be a support tool for quantum computing \cite{our_paper, chatgpt4_study}.

Other approaches exist to specialize LLMs to produce accurate answers for specific fields: for example, the plugin of wolframalpha for ChatGPT~\cite{chatGPT} promises ``powerful computation, accurate math, curated knowledge, real-time data and visualization'', or CrunchGPT~\cite{Kumar_2023} for scientific machine learning.
\chatforquant~is similar because it uses a pre-trained large language model and specialized modules. 
Its focus is however different as it targets specifically quantum computing.

We have recently come across two noteworthy new initiatives related to \chatforquant. 
The first one, Copilot in Azure Quantum \cite{chatbot_azure}, is developed by Microsoft as part of their learning platform. It introduces a ``quantum-focused chatbot designed to assist in code writing and enhance understanding of quantum concepts'' by utilizing Q\#. 
Copilot, built upon OpenAI's GPT-4, generates responseses through probabilistic computations, presenting a distinct approach compared to \chatforquant. 
The second initiative is the Quantum AI Chatbot \cite{GPT_quantum}, a customized version of ChatGPT claiming to be a ``Quantum Theory Translator and Progressive Trainer with Web Scraping.'' Similar to Copilot, this chatbot is also constructed upon OpenAI's GPT-4. While we have not yet examined the quality of these software solutions, a comprehensive evaluation is planned for future research.

Another approach \cite{chatbot_equations} involves a chatbot translating conversations into a system of equations, subsequently solved using a quantum circuit in Qiskit. However, the software itself is not currently available; only its methods and architecture have been described. Another relevant work is the chatbot developed by Vischnu Sivan \cite{chatbot_medium}, specializing in explaining quantum computing in layman's terms, though it does not encompass programming considerations.

%Evaluation of ease of learning quantum computing? (IS THERE ANYTHING?)

\section{Conclusions and Future Work} \label{s:conc}

The development of \chatforquant~ represents a new approach for developing chatbots. 
Instead of prioritizing human-like responsiveness, the development of \chatforquant~ focuses on answering correctly. 
\chatforquant~ is a reliable chatbot that adeptly responds simple inquiries about quantum computing. 
Prioritizing accuracy sets \chatforquant~ apart, since it acknowledges its limitations by indicating when it lacks the necessary information rather than providing incorrect or incomplete answers.
\chatforquant's reliability makes it a possible support tool for software developers, as well as for general users, who might not be experts in quantum computing.

Looking ahead, our aspirations for \chatforquant~ extend beyond its current capabilities. 
We envision enhancing its question-answering competencies by incorporating a broader range of topics, such as quantum cryptography and quantum random number generators. 
We also aim to integrate tasks that delve into the intricacies of software engineering. 
This will enable \chatforquant~ to provide insightful suggestions on software engineering challenges within the context of quantum computing. 
An additional objective is enabling \chatforquant~ not only to give answers but also to provide quantum code to users, thereby offering a more immersive experience.

On the way of continuous improvement, we ambition to refine the conversational dynamics of \chatforquant. 
We aspire to cultivate a more human-like responsiveness, taking into account the history of messages to enhance context awareness. 
Enabling users to ask multiple questions in a single interaction and incorporating a login feature to save conversations are additional features that we want to integrate. 
These enhancements are directed towards making the user experience more seamless and enriching.

%In summary, \chatforquant~ is a promising first step towards a tool for non-expert users to learn and employ quantum computing. 
%It is already able to accurately answer questions about quantum gates with an intuitive interface.

%Since the current performance of the QA-LLM module presents some slowness, we are particularly interested in investigating the performance of \chatforquant~ with other QA models.\newline

%Summarizing, \chatforquant~ has been marked by achievement, but our commitment to improvement drives us towards a more sophisticated and user-friendly chatbot for quantum computing.

\begin{acks}
The authors thank Ilgiz Mustafin for his help in the \chatforquant's deployment.
\end{acks}

\bibliographystyle{ACM-Reference-Format}
\bibliography{biblio}

\end{document}